\pdfoutput=1

\documentclass[11pt]{article}

\usepackage[preprint]{acl}

\usepackage{times}
\usepackage{latexsym}

\usepackage[T1]{fontenc}

\usepackage[utf8]{inputenc}

\usepackage{microtype}

\usepackage{inconsolata}

\usepackage{graphicx}
\usepackage{graphics}

\usepackage{amsmath}
\usepackage{booktabs}

\newcommand{\ie}{\textit{i.e.}}
\newcommand{\eg}{\textit{e.g.}}

\title{Resampling Benchmark for Efficient Comprehensive Evaluation of\\Large Vision-Language Models}

\author{Teppei Suzuki \\
  SB Intuitions \\
  \texttt{teppei.suzuki@sbintuitions.co.jp} \\\And
  Keisuke Ozawa \\
  SB Intuitions \\
  \texttt{keisuke.ozawa@sbintuitions.co.jp} \\}

\begin{document}
\maketitle
\begin{abstract}
We propose an efficient evaluation protocol for large vision-language models (VLMs).
Given their broad knowledge and reasoning capabilities, multiple benchmarks are needed for comprehensive assessment, making evaluation computationally expensive.
To improve efficiency, we construct a subset that yields results comparable to full benchmark evaluations.
Our benchmark classification experiments reveal that no single benchmark fully covers all challenges.
We then introduce a subset construction method using farthest point sampling (FPS).
Our experiments show that FPS-based benchmarks maintain a strong correlation (> 0.96) with full evaluations while using only ~1\% of the data.
Additionally, applying FPS to an existing benchmark improves correlation with overall evaluation results, suggesting its potential to reduce unintended dataset biases.

\end{abstract}

\begin{figure*}[t]
    \centering
    \includegraphics[width=0.8\linewidth]{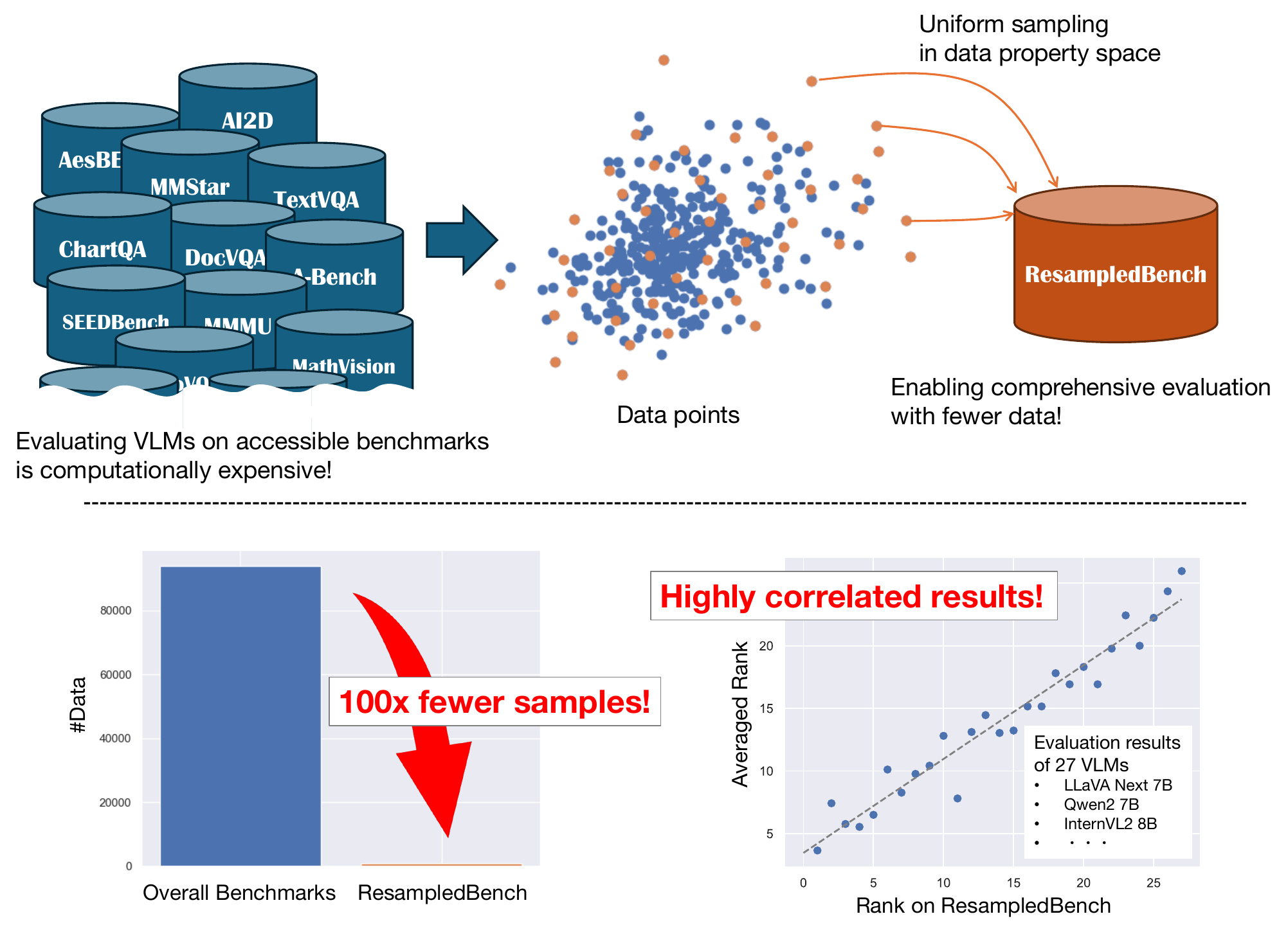}
    \caption{Since it is costly to evaluate VLMs on a large number of accessible benchmarks for the comprehensive evaluation, we sample data from the benchmarks with as fewer samples as possible to make evaluation efficient but as comprehensive as using all the data. As a result, our benchmark, ResampledBench, has approximately 100$\times$ fewer samples than sum of the number of data in the accessible benchmarks, while the model ranks on ResampledBench have a correlation coefficient of over 0.96 with the model ranks averaged over the ranks on all benchmarks.}
    \label{fig:overview}
\end{figure*}

\section{Introduction}
\label{sec:intro}

Large language models (LLMs) have significantly advanced machine intelligence~\cite{touvron2023llama,aizawa2024llm,team2024gemma,jiang2023mistral}.
Following this progress, large vision-language models (VLMs) have emerged~\cite{liu2024visual,instructblip,liu2024improved,bai2023qwen}, enabling performance across a wide range of image and language tasks.
The impressive capabilities and practicality of LLMs and VLMs have spurred industries and academic institutions to develop foundation models~\cite{you2023ferret,touvron2023llama,aizawa2024llm,team2024gemma,jiang2023mistral,bai2023qwen,alayrac2022flamingo}.

Model evaluation is a critical step in the development of VLMs.
Especially, since foundation models including VLMs are designed to support a large variety of tasks, diverse datasets and tasks are required to thoroughly assess their capabilities.
Numerous benchmarks have been developed to evaluate different aspects of these models' performance~\cite{zhang2024abench,zhang2024mathverse,wang2024measuring,gu2024mllmguard}.
Additionally, benchmarks covering comprehensive domains and tasks have been proposed to assess broader general knowledge and advanced reasoning abilities of VLMs~\cite{yue2023mmmu,li2023seed,li2023seed2,li2024seed2plus,MMBench,wang2024muirbench}.

However, there is currently no standardized policy for the collection and evaluation of these benchmarks.
In fact, VLMs are ranked differently depending on the benchmark used, indicating that model performance is biased towards specific data types and tasks.
This observation suggests that even the large benchmarks do not provide a comprehensive evaluation each alone.
To address this, an averaged ranking across multiple benchmarks is often employed to evaluate VLMs more comprehensively~\cite{duan2024vlmevalkit}. Averaging per-benchmark rankings is expected to give a more balanced assessment of models across various tasks and domains, beyond what any single benchmark can cover.
However, this approach requires costly evaluations using data from all benchmarks, which is resource-intensive. 

Given the recent increase in the number of benchmarks and the data therein, it is advantageous for both developers and users to speed up the comprehensive evaluation of VLMs.
Preparing a shortcut protocol that requires much less data but still allows us to approximate the performance trends of models as accurately as a full evaluation over the entire dataset would further benefit the community.

For the aforementioned reasons, in this work, we focus on improving the efficiency of evaluation.
We first conduct a dataset classification experiment~\cite{torralba2011unbiased} using 16 multiple choice question benchmarks~\cite{MMBench,li2023seed,li2023seed2,li2024seed2plus,lu2022learn,wu2023q,zhang2024abench,Kembhavi2016ADI,chen2024we,realworldqa,zhang2024task,schwenk2022okvqa,yue2023mmmu,chen2024gmai,wang2024divide,huang2024aesbench} and 11 visual question answering benchmarks~\cite{mishra2019ocr,singh2019towards,mathew2021docvqa,mathew2022infographicvqa,masry2022chartqa,hudson2019gqa,liu2023hidden,wang2024measuring,kim2024tablevqa,wang2024allseeing_v2,yu2023mm}. Through this experiment, we find that there is no single benchmark capable of enabling comprehensive evaluation, and also suggests that some overlap among specific benchmarks.
Based on these findings, we propose a protocol that constructs subsets from existing benchmarks using farthest point sampling (FPS) in the feature space, and then performs evaluations using these subsets.

We refer to the benchmark constructed through this sampling method as \textit{ResampledBench}.
It is significantly smaller in scale compared to the original benchmarks, thereby enabling more efficient evaluations.
In addition, we demonstrate that applying the proposed sampling strategy to MMStar~\cite{chen2024we}, which is carefully curated and balanced by human experts, improves its correlation with the average rank.
This result highlights its potential as a filtering technique in benchmark construction.
An overview of this study is presented in Fig. \ref{fig:overview}.

\section{Related Work}
\label{sec:related_work}
Evaluation is particularly important for continuous improvement and setting policies in machine learning community.
Especially, LLM-based models, including LLMs~\cite{touvron2023llama,aizawa2024llm,team2024gemma,jiang2023mistral}, VLMs~\cite{liu2024visual,instructblip,liu2024improved,bai2023qwen}, have broader knowledge and advanced reasoning capabilities, and they should be evaluated on multiple benchmarks for assessing their capabilities from diverse viewpoints~\cite{kwiatkowski2019natural,liu2024visual,Li-hallucination-2023,hendrycks2020measuring,clark2019boolq,lin2021truthfulqa,wang2024muirbench,tanaka2023slidevqa,luo2024jailbreakv}.
To make evaluation of VLMs feasible comprehensively on multiple benchmarks, some platforms have been developed evaluation toolkits on multiple benchmarks with the same manner~\cite{zhang2024lmms,duan2024vlmevalkit}.

\begin{table*}[t]
    \centering
    \caption{Classification accuracy on multiple choice question (MCQ) task benchmarks for various input data types. I, Q, and A of input data denote image, question, and answer, respectively. \label{tab:mcq}}
    \scalebox{0.85}{
    \begin{tabular}{lccccccc}
        \toprule
        Input & I & Q & A & I~+~Q & I~+~A & Q~+~A & I~+~Q~+~A \\ \midrule
        Acc. & 0.752$\pm$0.007 & 0.540$\pm$0.013 & 0.105$\pm$0.018 & 0.793$\pm$0.010 & 0.762$\pm$0.005 & 0.540$\pm$0.014 & 0.791$\pm$0.014 \\ \bottomrule
    \end{tabular}
    }
\end{table*}
\begin{table*}[t]
    \centering
    \caption{Classification accuracy on visual question answering (VQA) task benchmarks for various input data types. I, Q, and A denote image, question, and answer, respectively. \label{tab:vqa}}
    \scalebox{0.85}{
    \begin{tabular}{lccccccc}
        \toprule
        Input & I & Q & A & I~+~Q & I~+~A & Q~+~A & I~+~Q~+~A \\ \midrule
        Acc. & 0.757$\pm$0.012 & 0.900$\pm$0.005 & 0.893$\pm$0.003 & 0.983$\pm$0.003 & 0.993$\pm$0.003 & 0.958$\pm$0.005 & 0.993$\pm$0.002 \\ \bottomrule
    \end{tabular}
    }
\end{table*}

A wide range of benchmarks are currently used for VLM evaluation. 
Most of these benchmarks focus on assessing certain capabilities, such as text-rich images (\eg, documents, charts, diagrams, and infographics~\cite{mishra2019ocr,liu2023hidden,li2024seed2plus,mathew2022infographicvqa}), low-level vision tasks~\cite{rahmanzadehgervi2024vision,wu2023q}, and subject knowledge at high school and college levels~\cite{yue2023mmmu,lu2022learn}.

While there are benchmarks aimed at comprehensive evaluation~\cite{zhang2024lmms,chen2024we,li2023seed,MMBench,ying2024mmt}, achieving a truly exhaustive assessment of VLMs still requires the use of multiple benchmarks.
This is because new challenges and evaluation perspectives continually emerge, necessitating a broader set of evaluation criteria. 
For example, HallusionBench~\cite{guan2023hallusionbench} was introduced to address hallucination issues, and BlindTest~\cite{rahmanzadehgervi2024vision} was proposed in response to findings that VLMs struggle with low-level vision tasks, such as counting line intersections or identifying rows and columns in grids. 
These benchmarks were developed from newly emerging perspectives as VLM evaluation techniques advanced, making it challenging to anticipate such specific evaluation needs from the outset.

Evaluating a model based on its aggregated performance across multiple benchmarks provides a more reliable measure of its overall capabilities compared to using a single benchmark.
However, performing evaluations on multiple benchmarks is computationally expensive and resource-intensive.
There are some studies for making LLM evaluation on multiple benchmarks more efficient~\cite{polo2024tinybenchmarks,perlitz2023efficient}, but their procedure cannot be directly applied to multimodal data.
Thus, we analyze data for VLM evaluation from the dataset classification perspective, and propose a simple procedure for efficient evaluation.

\begin{figure*}[t]
    \centering
    \begin{tabular}{cc}
        \includegraphics[width=0.4\linewidth]{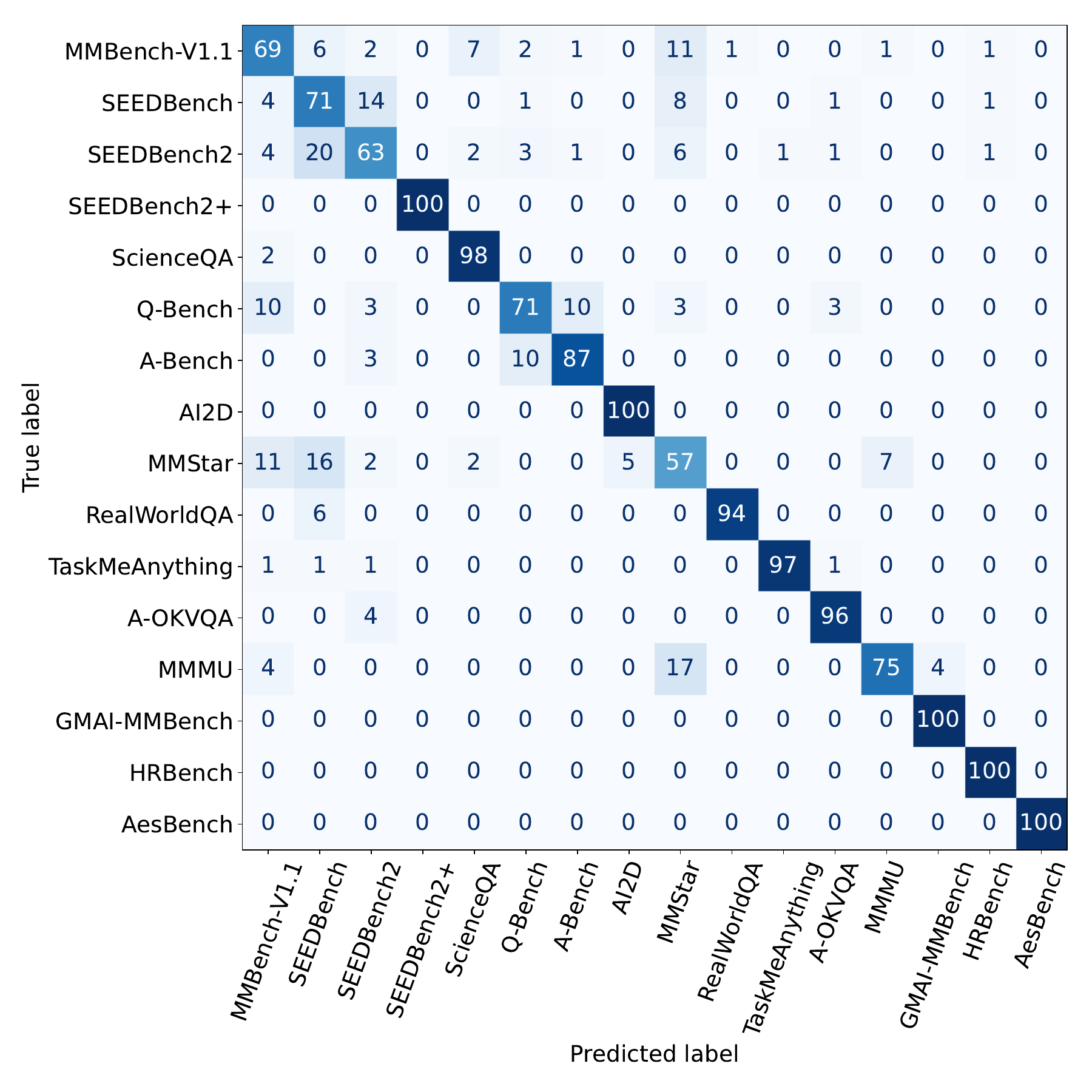} &
        \includegraphics[width=0.4\linewidth]{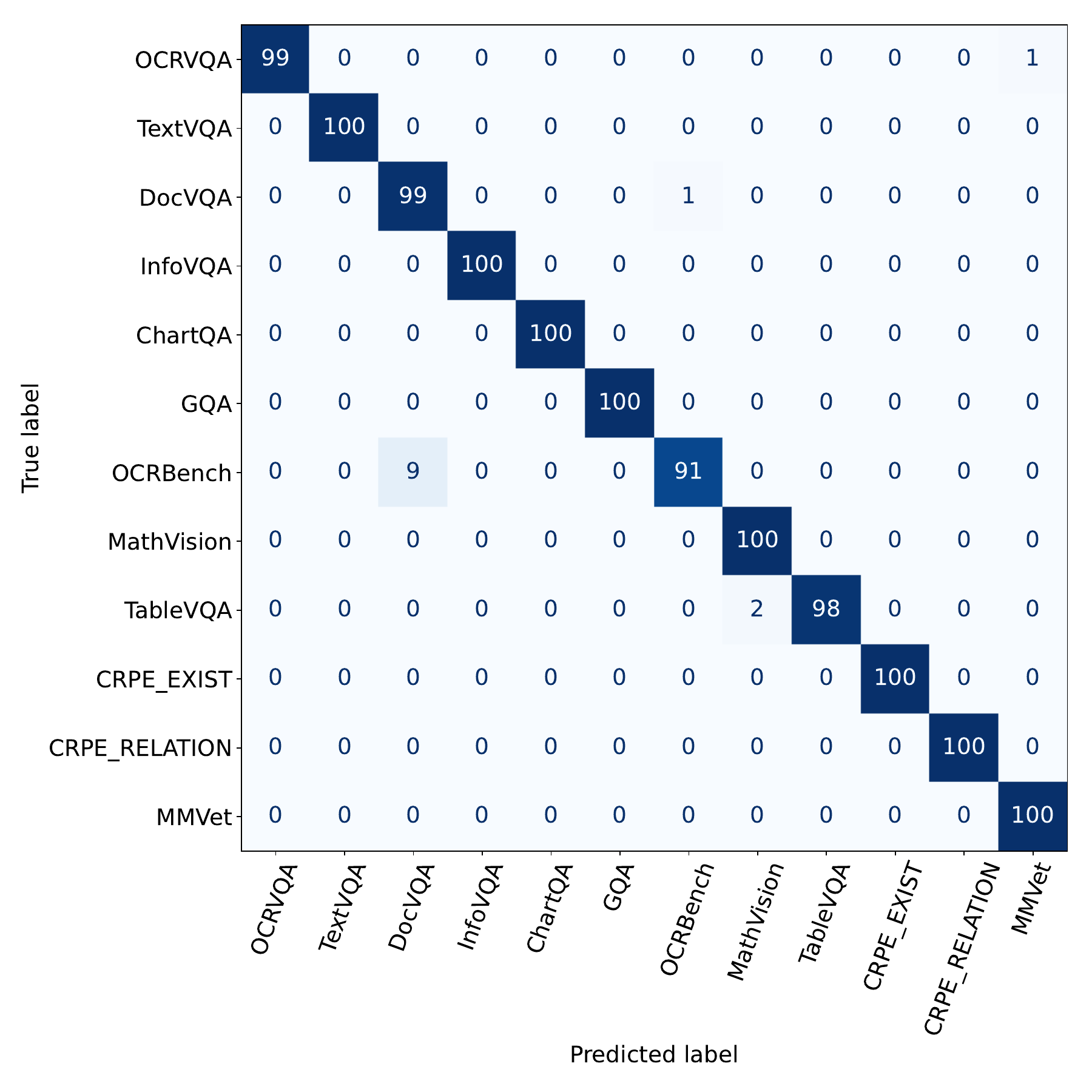} \\
        (a) MCQ &
        (b) VQA
    \end{tabular}
    \caption{Confusion matrices of the benchmark classification experiments for MCQ with image and question input and VQA with image, question, and answer input.}
    \label{fig:conf-mat}
\end{figure*}

\section{Dataset Classification for Assessing Comprehensiveness}
\label{sec:bench-cls}
In this section, we conduct the dataset classification experiment~\cite{torralba2011unbiased} to verify how much overlap exists in the tasks and domain data in each benchmark.
Intuitively, data in a benchmark collected comprehensively over diverse domains and tasks should be misclassified as belonging to the other benchmarks, because similar data would be found in the other benchmarks.
Such a benchmark can therefore serve as a comprehensive evaluator.

We randomly sample 1,000 data points as training data from each benchmark or 80\% samples of the available data if the benchmark contains fewer than 1,000 samples, with remaining 20\% samples used as the test data.
We train a dataset classifier consisting of a linear classifier on top of a pretrained encoder using the linear probing protocol, meaning that only the linear classifier is trained.

We focus on two major task formats for VLM evaluation: multiple choice questions (MCQ), where an answer is selected from a set of options, and visual question answering (VQA), which requires generating open-ended answers.
We select 16~\cite{MMBench,li2023seed,li2023seed2,li2024seed2plus,lu2022learn,wu2023q,zhang2024abench,Kembhavi2016ADI,chen2024we,realworldqa,zhang2024task,schwenk2022okvqa,yue2023mmmu,chen2024gmai,wang2024divide,huang2024aesbench} for MCQ and 11 benchmarks~\cite{mishra2019ocr,singh2019towards,mathew2021docvqa,mathew2022infographicvqa,masry2022chartqa,hudson2019gqa,liu2023hidden,wang2024measuring,kim2024tablevqa,wang2024allseeing_v2,yu2023mm} for VQA.
Note that we do not use data from original sources but instead rely on VLMEvalKit~\cite{duan2024vlmevalkit}, as it provides standardized data formats across benchmarks, and the categorization of benchmarks (\ie, MCQ or VQA) follows the VLMEvalKit guidelines.

Since VLM benchmarks commonly consist of three data types (\ie, images, questions, and answers), we conduct the dataset classification using various input combinations, such as images only, answer only, and a combination of images and questions.
For text embedding, we use pretrained DeBERTa-v3-base~\cite{he2021debertav3}, and for image embedding, we use CLIP ViT-L/14~\cite{clip,dosovitskiy2020image}.
After independently encoding the texts and images, we concatenate the resulting text and image embeddings and input them into the linear classifier.

We train the linear classifier for 10 epochs with a batch size of 64.
The parameters are updated using the Adam optimizer with the default hyperparameters~\cite{KingBa15}.

The classification accuracy for each input is shown in Tabs. \ref{tab:mcq} and \ref{tab:vqa}.
We report averaged accuracy and standard deviation over 5 trials.
Except for the MCQ task using answer inputs, the model performs well across all conditions.
We attribute that these results are mainly due to the selection biases.
Most benchmarks are intentionally collected from specific domains and tasks, such as scientific domains~\cite{yue2023mmmu,Kembhavi2016ADI,lu2022learn}, OCR tasks~\cite{mathew2021docvqa,liu2023hidden,kim2024tablevqa,mishra2019ocr}, and mathematics~\cite{wang2024measuring}.
In addition, benchmarks focusing on the similar domains and tasks, such as OCRVQA~\cite{mishra2019ocr} and OCRBench~\cite{liu2023hidden}), are highly discriminative, which may be due to unintended selection bias.
These results suggest that each benchmark evaluates specific domains and tasks, and there is no benchmark that can comprehensively assess the full capabilities of VLMs.

The lower accuracy observed for MCQ using answer inputs is due to the fact that the answers are typically symbolic and lack meaningful content.
As a result, the inclusion of answers does not improve accuracy when comparing the results of using only images and questions versus using images, questions, and answers together.
In contrast, for VQA, the answers are more informative than the questions, as they describe the content of the input image, and the questions are somewhat uninformative.
Therefore, adding questions does not enhance accuracy when combined with image and answer inputs.

We also present confusion matrices for the MCQ task using image and question inputs, and for the VQA task using image, question, and answer inputs in Fig. \ref{fig:conf-mat}.
The confusion matrices for the other configurations are available in the appendix.

As shown in Tab. \ref{tab:vqa}, VQA benchmarks are classified almost perfectly.
That is because most VQA benchmarks focus on specific tasks or domains, such as an OCR task and document understanding, which simplifies the classification of VQA benchmarks.
Some data from OCRBench~\cite{liu2023hidden} is misclassified as DocVQA~\cite{mathew2021docvqa}, which can be attributed to the fact that 50 out of the 1,000 samples in OCRBench are derived from DocVQA and some of these samples are included in the randomly split test data.

MMStar~\cite{chen2024we} is the most challenging benchmark to classify.
It is composed of carefully balanced and purified samples, and contains various domain images, encompassing a wide range of domains such as illustrations, natural images, charts, and art.
Additionally, each selected sample is verified by human reviewers to ensure visual dependency, minimal data leakage, and the need for advanced multimodal capabilities for problem-solving
These aspects contribute to MMStar's comprehensive.
Nevertheless, more than half of the data could be classified correctly, which is above the chance rate.
This suggests that over half of MMStar's data is distinct when compared to other benchmarks.

Other hard-to-classify benchmarks, such as MMBench~\cite{MMBench} and SEEDBench~\cite{li2023seed}, are also designed to be comprehensive.
Conversely, the benchmarks that are easier to classify, such as SEEDBench2+~\cite{li2024seed2plus} and AI2D~\cite{Kembhavi2016ADI}, are biased toward specific fields and domains, such as scientific fields and chart images.

From another perspective, some data remains misclassified, indicating that there is overlap in the types of data included across benchmarks.
In other words, evaluating multiple benchmarks may involve some degree of redundant assessment.
Therefore, in the next section, we explore ways to optimize the evaluation process by reducing this redundancy.

In summary of the insights from the experiments in this section, there are no benchmarks capable of comprehensively assessing VLM capabilities.
To evaluate VLMs from diverse perspectives, multiple benchmarks are necessary. 
However, evaluating dozens of benchmarks is time-consuming and somewhat redundant, as evidenced in Fig. \ref{fig:conf-mat}, where some benchmarks cannot be classified with 100\% accuracy, suggesting that some data overlaps.

\section{Ranking Benchmarks and Resampling as Comprehensive evaluators}
\label{sec:rank-bench-resample}
\subsection{Ranking Benchmarks}
\label{sec:rank-bench}
The fact that the benchmarks can be classified with high accuracy rates suggests that certain biases are present in these benchmarks. Additionally, each model’s rank on individual benchmarks may be affected by benchmark-specific biases, such as data collection policies and human preferences.
To address this issue, we use averaged ranks to mitigate such biased evaluations by averaging all ranks measured across all accessible benchmarks.

Given a set of models $\mathcal{M} = \{M_i\}_{i=1,2,\cdots}$ to be evaluated and a set of benchmarks $\mathcal{B} = \{B_j\}_{j=1,2,\cdots}$, where $B_j$ consists of triplets $\{(I_k,Q_k,A_k)\}_{k=1,2,\cdots,|B_j|}$ of an image $I_k$, a question $Q_k$, and an answer $A_k$, the averaged rank of model $M_i$ is defined as follows:

{\small
\begin{align}
    \label{eq:avg_rank}
    \mathrm{AvgRank}(M_i|\mathcal{M}, \mathcal{B}) = \frac{1}{|\mathcal{B}|}\sum_{B_j\in\mathcal{B}}\mathrm{Rank}(M_i|\mathcal{M},B_j),
\end{align}
}%
where $|\mathcal{B}|$ is the total number of the benchmarks, and the ranks of model $M_i$ on each benchmark $\mathrm{Rank}(M_i|\mathcal{M},B_j)$ is determined by its score relative to the other models $\{ (M_{i'}) \}_{i' \neq i}$.
In the experiments, we compute the averaged ranks of 27 models using 26 benchmarks, which are listed in Appendix. 

We consider $\mathrm{AvgRank}$ to represent a comprehensive evaluation of a model's capabilities.
Based on this assumption, we evaluate the comprehensiveness of each benchmark by measuring the correlation between the ranks on individual benchmarks and $\mathrm{AvgRank}$.
For the correlation between ranks, we evaluate using Spearman's rank correlation coefficient in our experiments.

\subsection{Resampling Benchmarks for efficient and comprehensive evaluation}
\label{sec:resample-bench}
$\mathrm{AvgRank}$ is the result of evaluating the models in a comprehensive manner, as opposed to evaluating them on individual benchmarks.
Unfortunately, the total number of samples in the benchmarks can often be quite large, which incurs significant costs in terms of both time and computational resources.
To reduce the evaluation cost while maintaining the comprehensiveness of models assessments, we resample data from the accessible benchmarks.
The goal is to use as few samples as possible to make the evaluation more efficient, yet still achieve a level of comprehensiveness comparable to using the full set of data.

To achieve this, we consider sampling data from the existing datasets such that the sampling is uniformly distributed across a space where data properties, such as tasks and domains, are well-described.
Based on the results of the benchmark classification experiments conducted in the previous section, we found that the benchmarks can be effectively classified within the feature space spanned by the pretained encoders.
This suggests that the feature space constructed by pretrained models captures the differences between tasks and domains.
Thus, we assume that this feature space can serve for these properties and proceed with evaluations within this space.
 
The sampling strategy is crucial for ensuring that the sampled data comprehensively covers the entire dataset. 
To enhance comprehensiveness of the benchmark, data should be sampled uniformly in the data property space.
For this purpose, we use farthest point sampling (FPS)~\cite{eldar1997farthest}.
It is widely used for point cloud sampling in 3D computer vision~\cite{qi2017pointnet++,qi2019deep,yan2020pointasnl}, and we adopt it to sample points uniformly across the spatial dimensions of the feature space.

Let $\mathcal{X}=\{x_i\}_{i=1,2,3,\dots}$ be a set of data points and $f$ be an image-text encoder consisting of DeBERTa-v3-base~\cite{he2021debertav3} and CLIP ViT-14/L~\cite{clip,openclip}.
Define $\mathcal{S}_0$ as a set consisting of a query data point randomly selected from $\mathcal{X}$.
Then, an $(S)$-th sampled point by FPS is defined as follows:

{\footnotesize
\begin{align}
    \mathcal{S}_{S}=\mathcal{S}_{S-1}\cup\left\{\underset{x_i\in \mathcal{X}\setminus\mathcal{S}_{S-1}}{\arg\max}\left[\min_{x_j\in \mathcal{S}_{S-1}}D(f(x_i),f(x_j))\right]\right\},
\end{align}
}%
where $D(\cdot,\cdot)$ is an arbitrary distance function, and we use the Euclidean distance in our experiments.


We believe this approach not only samples diverse data, but is also expected to reduce biases within the dataset because of the uniformity induced by FPS.
Furthermore, our experiments suggest that applying FPS in the feature space can sever as a filtering process to mitigate the dataset biases.

\section{Experimental Results}
\label{sec:recons-bench}

We evaluate the comprehensiveness of our resampled benchmarks, ResampledBench, by examining whether the ranks measured on them show high correlations with the averaged ranking ($\mathrm{AvgRank}$).
To assess how effectively existing benchmarks serve as comprehensive evaluators and whether our resampled benchmarks can provide a better evaluation, we compare the Spearman correlations of the existing benchmarks and our resampled ones with $\mathrm{AvgRank}$.

In all the experiments, we evaluate 27 VLMs with roughly 1B to 10B parameters~\cite{chen2023internvl,chen2024far,bai2023qwen,wang2024qwen2,llava-next,laurenccon2024obelics,li2023monkey,huang2024mini,yao2024minicpm,lu2024ovis,abdin2024phi,wemm,internlmxcomposer2,glm-4v,h2ovl,blip3-xgenmm} on these benchmarks for computing the ranks because most publicly available VLMs have fewer than 15 billion parameters as shown in Fig. \ref{fig:vlm-params} in the appendix.
The models are listed in the appendix.
We select 26 benchmarks~\cite{MMBench,li2023seed,li2024seed2plus,lu2022learn,wu2023q,zhang2024abench,Kembhavi2016ADI,chen2024we,realworldqa,zhang2024task,schwenk2022okvqa,yue2023mmmu,chen2024gmai,huang2024aesbench,mishra2019ocr,singh2019towards,mathew2021docvqa,mathew2022infographicvqa,masry2022chartqa,hudson2019gqa,liu2023hidden,wang2024measuring,wang2024allseeing_v2,yu2023mm} from Fig. \ref{fig:conf-mat}, each data of which consists of a single image with a question written in English, and the answers are provided in text.
Benchmarks requiring multiple inputs are omitted in our evaluation as there are relatively few models that support multiple languages, multiple image inputs, or image outputs.
For example, SEEDBench2~\cite{li2023seed2} requires image generation skills to answer the questions.
The total number of data points across all the benchmarks is 93,996.
We follow the VLMEvalKit evaluation protocol~\cite{duan2024vlmevalkit}.

\begin{figure}
    \centering
    \includegraphics[width=1\linewidth]{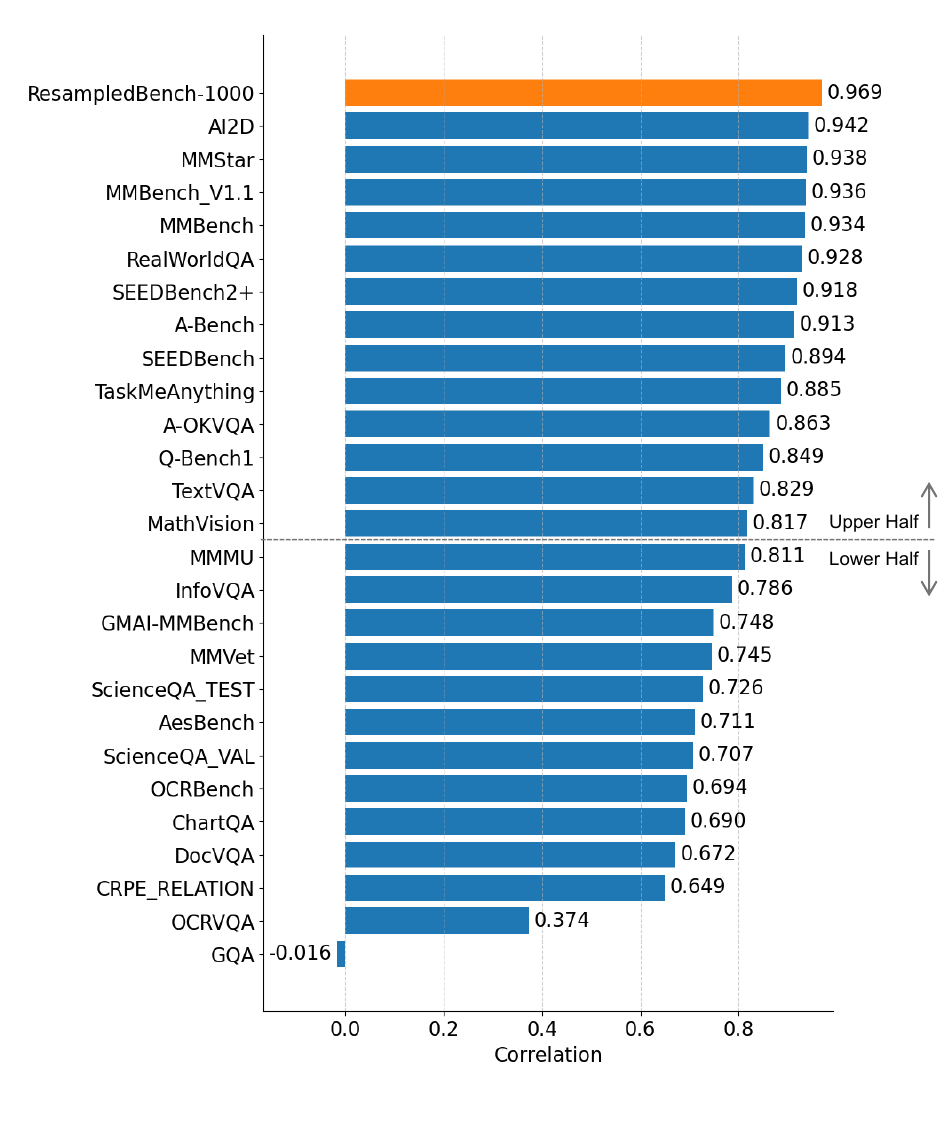}
    \caption{The correlations between $\mathrm{AvgRank}$ and the ranks on individual benchmarks.}
    \label{fig:corr_rank}
\end{figure}

\subsection{How Comprehensive Existing Benchmarks Are}
We evaluate the correlations between $\mathrm{AvgRank}$ and the ranks on individual benchmarks.
Fig. \ref{fig:corr_rank} shows the ordering of the correlations.
Benchmarks such as MMStar~\cite{chen2024we}, MMBench series~\cite{MMBench}, SeedBench series~\cite{li2023seed,li2024seed2plus} show relatively high correlations with $\mathrm{AvgRank}$.
These benchmarks include data from various categories and cover diverse tasks to measure reasoning capabilities of VLMs.
In contrast, ChartQA~\cite{masry2022chartqa}, DocVQA~\cite{mathew2021docvqa}, CRPE\_RELATION~\cite{wang2024allseeing_v2}, and OCRVQA~\cite{mishra2019ocr} show lower correlations, which aligns with the fact that these benchmarks are designed to evaluate specific skills.
Among those with lower correlations, GQA~\cite{hudson2019gqa} shows particulary low correlation.
This is considerably because this benchmark focuses on assessing compositionality understanding and has inherent biases when compred to the others.
We also see that MMStar~\cite{chen2024we}, which ranks second in correlation, focuses on a comprehensive evaluation and indeed shows a strong correlation with $\mathrm{AvgRank}$.
This result supports our hypothesis that using $\mathrm{AvgRank}$ for comprehensive evaluation aligns with the tendencies observed in the existing benchmarks.
Notably, AI2D~\cite{Kembhavi2016ADI} has the highest correlation despite its focus on scientific diagrams.
This is likely because AI2D~\cite{Kembhavi2016ADI} compensates for the bias toward the image domain by requiring diverse capability, such as OCR skills, complex reasoning, and general knowledge in addition to the scientific knowledge.

\subsection{Evaluating the Comprehensiveness of ResampledBench}
In this section, we verify that our resampled benchmarks enable comprehensive evaluation.
As a baseline approach, we evaluate a random sampling strategy from benchmarks.
In the random sampling, the probability of sampling data from a specific benchmark is proportional to the number of data points within that dataset.
This sampling strategy is reasonable because it aligns with the benchmark composition ratios in calculating $\mathrm{AvgRank}$.

\begin{figure}[t]
    \centering
    \includegraphics[width=0.9\linewidth]{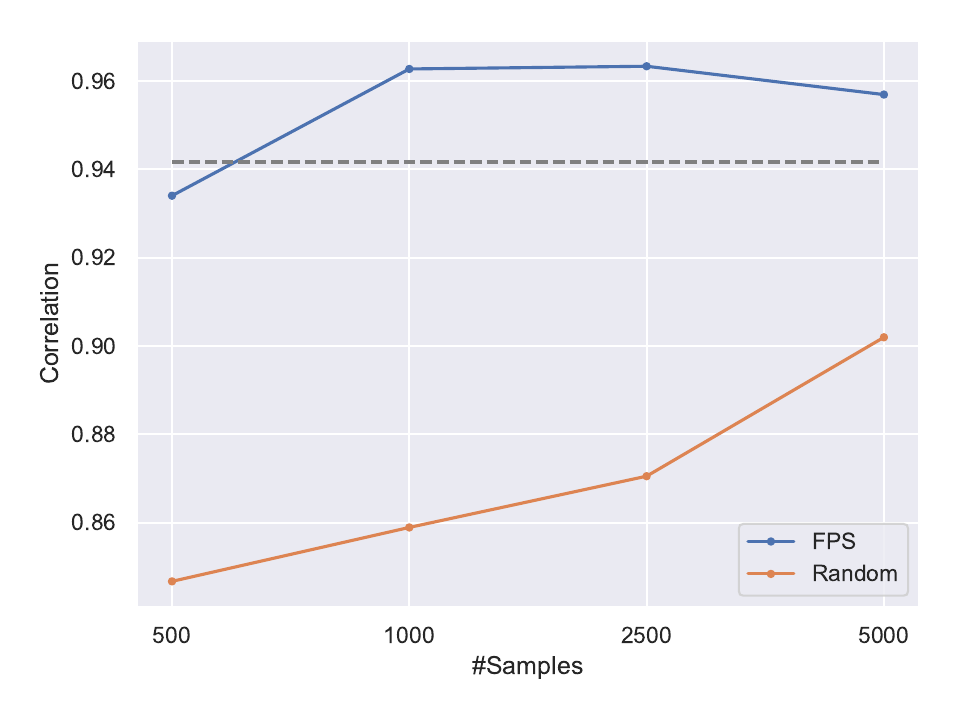}
    \caption{The correlation coefficient of a various number of data. The broken line denotes the correlation coefficient of AI2D that has the highest correlation coefficient among the existing benchmarks listed in Fig. \ref{fig:corr_rank}. Compered to random sampling, FPS achieves higher correlation even with 10$\times$ fewer samples.}
    \label{fig:fps_vs_random}
\end{figure}

\begin{table}[t]
    \centering
    \caption{Correlations of 1000 sampled benchmarks using different resampling strategies to $\mathrm{AvgRank}$. We report the mean and standard deviation over 3 trials.}  
    \label{tab:UpperLowerAll_1000}
    \scalebox{0.8}{
    \begin{tabular}{cccc}
        \toprule
        Benchmark      & Upper Half & Lower Half & All \\ \midrule
        Random   & 0.964$\pm$0.009 & 0.701$\pm$0.015 & 0.836$\pm$0.026 \\
        FPS      & 0.969$\pm$0.002 & 0.726$\pm$0.019 & 0.969$\pm$0.006 \\ \bottomrule
    \end{tabular}
    }
\end{table}

Fig. \ref{fig:fps_vs_random} compares the ranking correlations of FPS- and Random-resampled benchmarks with different number of samples.
FPS exhibits higher correlations to $\mathrm{AvgRank}$ than Random for all the four numbers of sampling.
This result indicates that using FPS provides more comprehensive benchmarks than random sampling from the whole datasets. 
Also, on top of Fig. \ref{fig:corr_rank}, we place our ResampledBench-1000 sampled 1000 points using FPS, and we have the higher correlation than all the other benchmarks.
AI2D, which shows the second highest correlation in Fig. \ref{fig:corr_rank}, contains 3,088 data points, which is three times the amount used by our proposed method.
The 1,000 samples are roughly 100 $\times$ fewer than the total number of data points across all benchmarks.
The computational speed cannot be definitively stated, as it depends on the parallelism of the hardware and the model itself.
However, ResampleBench-1000 can, in simple terms, evaluate models with 100 $\times$ greater efficiency.

We observe that the accuracy plateaued at around 1,000 samples.
This suggests that approximately 1,000 samples are sufficient to comprehensively evaluate the tasks covered by the benchmarks used in this experiment.
Note that, as shown in eq. \eqref{eq:avg_rank}, $\mathrm{AvgRank}$ is a continuous value, whereas the rank calculated on a single benchmark is an integer. Therefore, the correlation coefficient between them will not be exactly 1.
When sampling 5,000 data by FPS, the correlation is a little lower than that with 1,000 and 2,500 data.
This is most likely caused by a distortion in the feature space.
In Sec. \ref{sec:bench-cls}, we confirmed that the feature space of the pretrained model can separate differences between benchmarks; however, this does not guarantee that the feature space captures the finer characteristics of the data.
We believe that with a small number of samples, it is possible to achieve uniform sampling by FPS that follows the major differences in tasks and domains.
However, as the number of samples increases, the sampling process starts to pick up distortions in the feature space, leading to biases in the data.
As a result, this may have caused the drop in correlation observed with 5,000 samples.

We also evaluate FPS and random sampling with different data sources to see their tendencies under various scenarios given more or less diverse benchmarks.
Based on Fig. \ref{fig:corr_rank}, we divide the existing benchmarks into two groups based on the level of correlation.
We then sample from the high-correlation group (upper half) and the low-correlation group (lower half) separately.

As shown in Tab. \ref{tab:UpperLowerAll_1000}, FPS shows higher correlation than random sampling across all settings.
In addition, FPS achieves the same correlation under sampling from the upper half and from all benchmarks, which implies that FPS can stably sample diverse data even as the number of benchmarks increases.
For the sampling from the lower half set, FPS cannot achieve higher correlation than the highest correlation among the benchmarks in the lower half set.
The benchmarks in the lower half set have relatively more bias.
We believe that such a biased data distribution amplifies the impact of distortions in the feature space during sampling.

\subsection{Toward Automated Filtering of Benchmarks with Unintended Bias}
The results so far suggest that FPS can uniformly sample a diverse range of data, and to further confirm this, we will investigate whether sampling a subset of data from the existing benchmark, MMStar~\cite{chen2024we}, can mitigate unintended biases.
MMStar is carefully balanced and purified by human experts and is designed to benchmark 6 core capabilities and 18 detailed axes.
Since, as discussed in \cite{torralba2011unbiased}, it is difficult to completely eliminate selection bias, there may be unintended biases.
Thus, by sampling data from MMStar with FPS as filtering and evaluate correlation to $\mathrm{AvgRank}$, we confirm FPS on the feature space uniformly samples diverse data.

MMStar contained 1,500 samples and we sample 1,000 samples by FPS.
We perform this sampling three times with different random seeds and evaluate the mean and standard deviation of the correlation with $\mathrm{AvgRank}$.

The correlation between $\mathrm{AvgRank}$ and the sampled benchmark is $0.956\pm0.002$, which is higher than the correlation between MMStar and $\mathrm{AvgRank}$ as shown in Fig. \ref{fig:corr_rank} (0.938).
This indicates FPS on the feature space can mitigate the unintended biases contained in the benchmark, and highlights its potential for use as a filtering technique in benchmark construction.

\section{Conclusion}
We addressed the challenge of efficiently evaluating large vision-language models (VLMs), which are increasingly important in industry and academia.
Our dataset classification experiments revealed overlaps among benchmarks, highlighting the lack of a single comprehensive evaluation benchmark.
To address this, we proposed a protocol using farthest point sampling (FPS) in feature space to construct subsets from existing benchmarks.
This method enables efficient evaluation using a fraction of the data while maintaining a strong correlation (> 0.96) with full benchmark assessments.

Experiments using FPS on MMStar demonstrate its potential for reducing bias in benchmark construction.
As future work, we aim to extend this approach to create new, more diverse, and unbiased benchmarks, enhancing VLM evaluation.



\textbf{Limitation.}
As seen in Fig. \ref{fig:fps_vs_random}, our sampling strategy slightly degraded correlation when the number of samples increased.
This indicates that the feature space is somewhat distorted, causing biases to emerge as the sample size increases, even if the sampling is spatially uniform.
The distortion in the feature space is likely due to the fact that the text and image encoders were trained independently.
Therefore, as a future direction, we aim to work on learning a multimodal embedding space that integrates both text and image information.

\bibliography{custom}

\appendix


\section{Additional Results of Benchmark Classification}
\begin{figure*}
    \centering
    \begin{tabular}{cc}
         \includegraphics[width=0.4\linewidth]{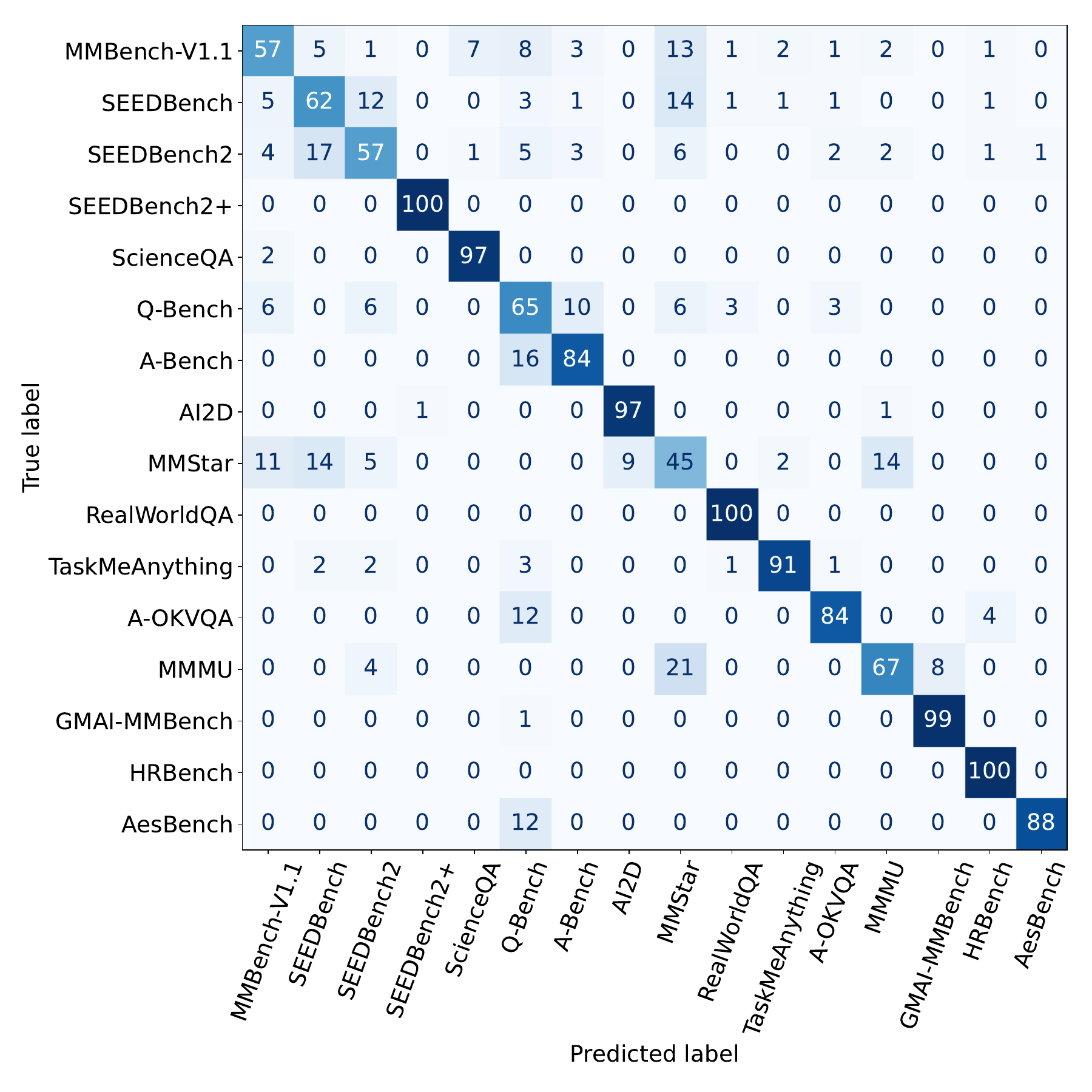} &
         \includegraphics[width=0.4\linewidth]{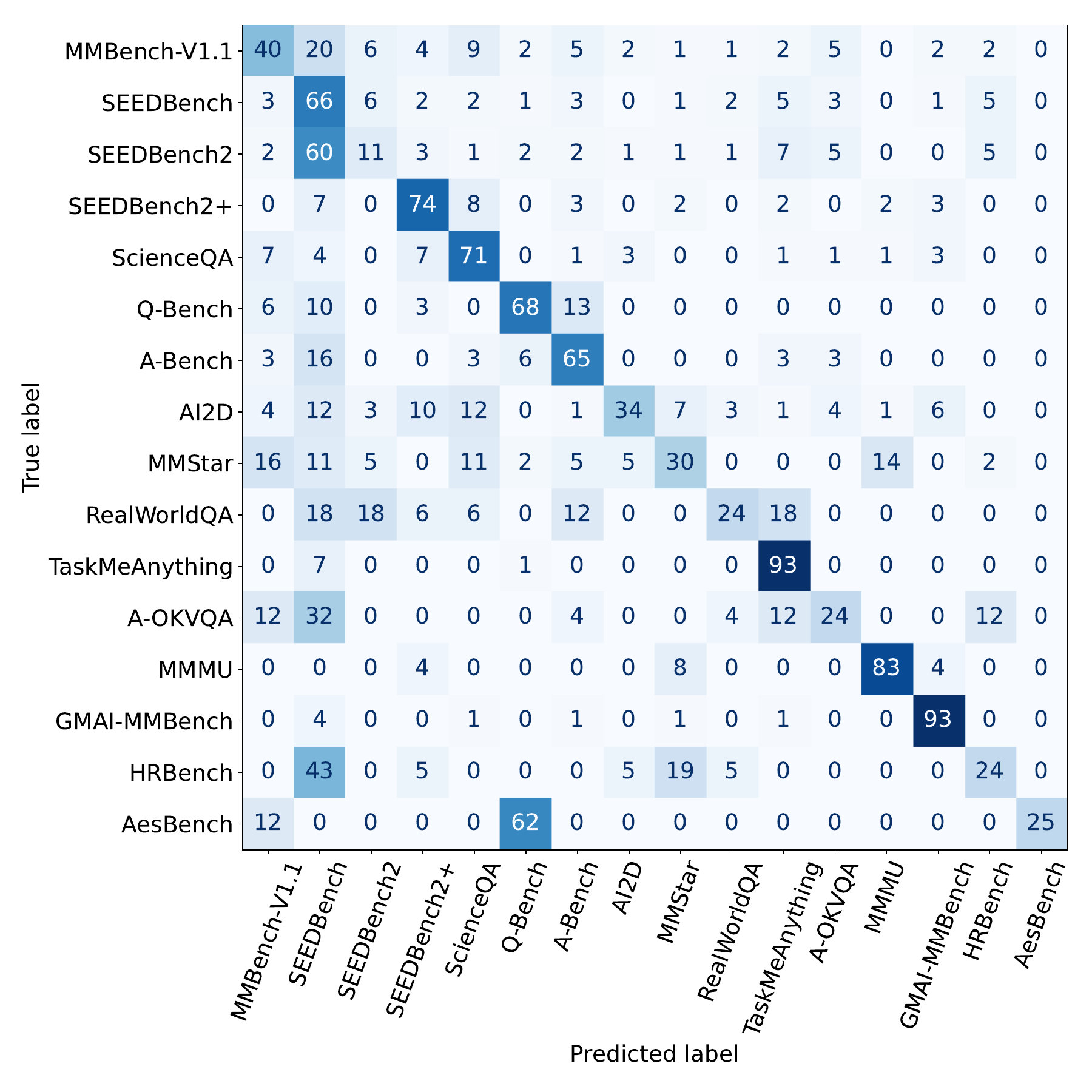} \\
         (a) Image & (b) Question \\
         \includegraphics[width=0.4\linewidth]{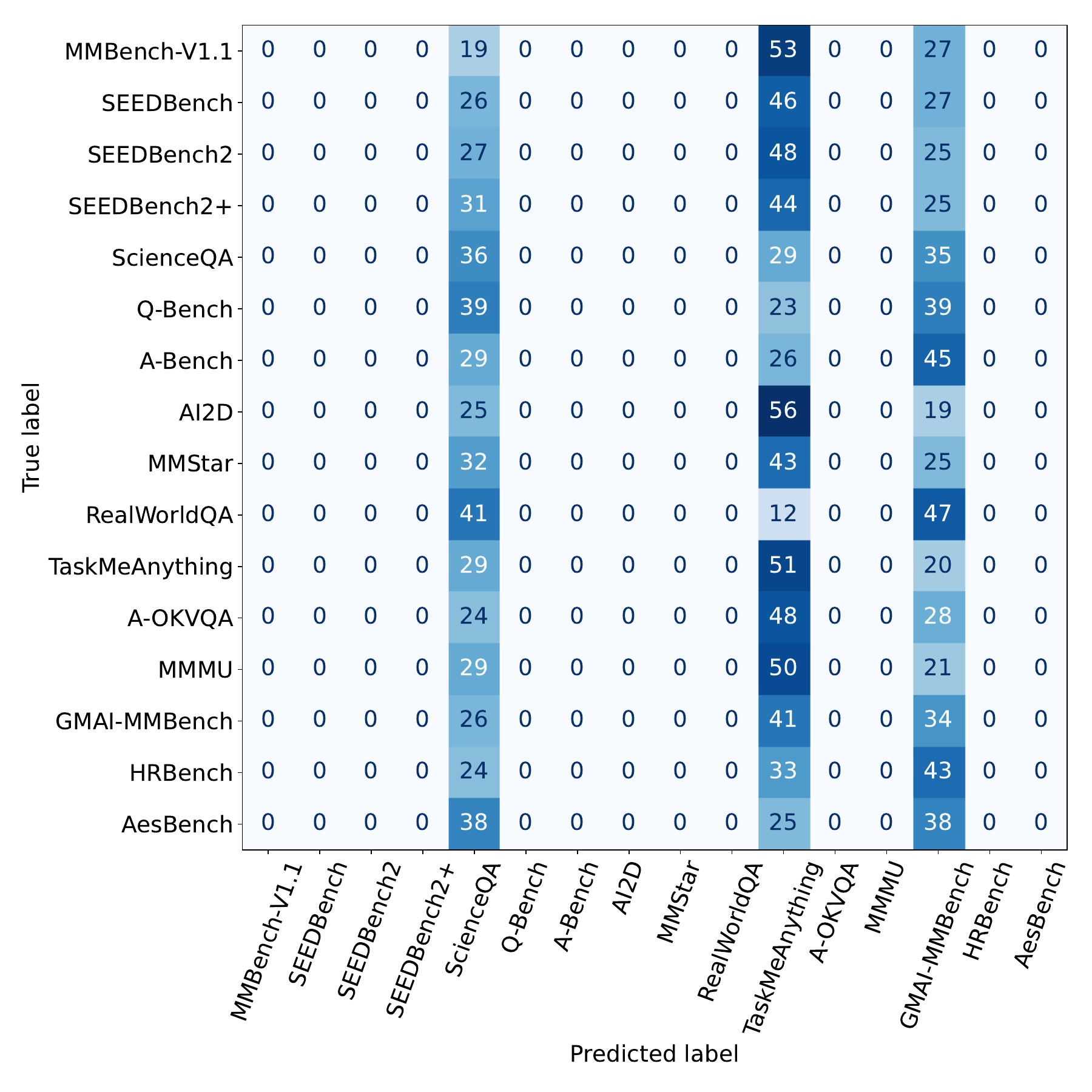} &
         \includegraphics[width=0.4\linewidth]{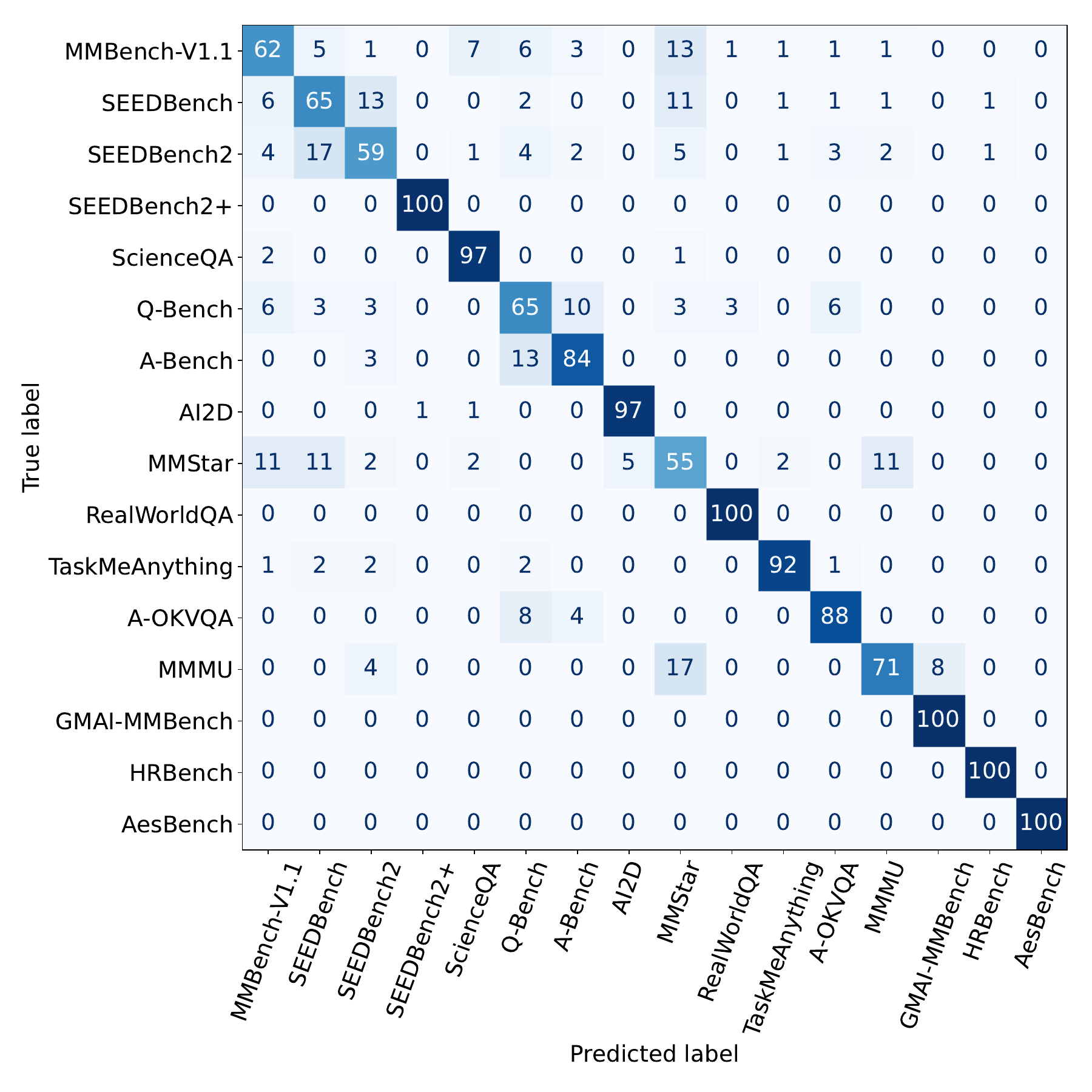} \\
         (c) Answer & (d) Image + Answer \\
         \includegraphics[width=0.4\linewidth]{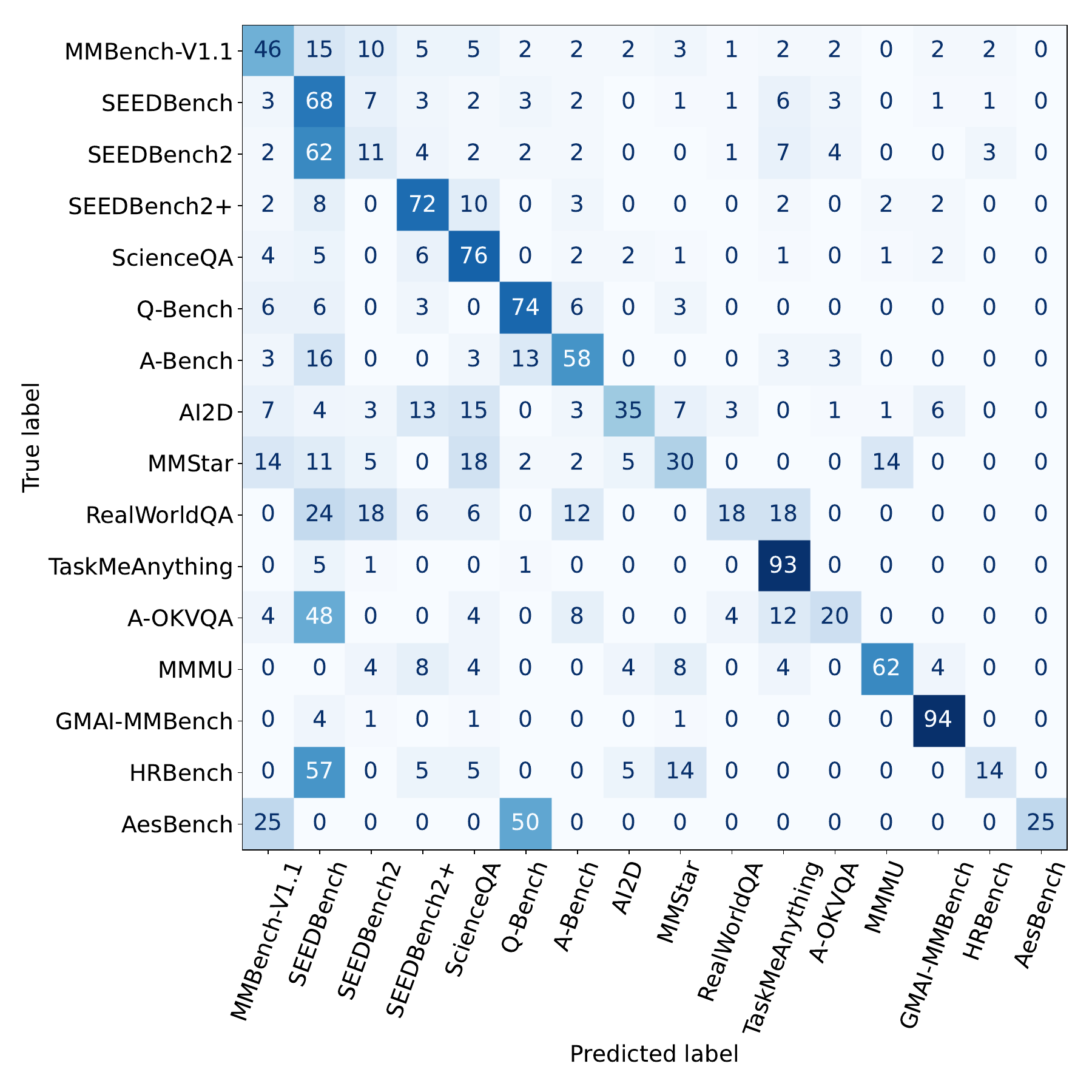} &
         \includegraphics[width=0.4\linewidth]{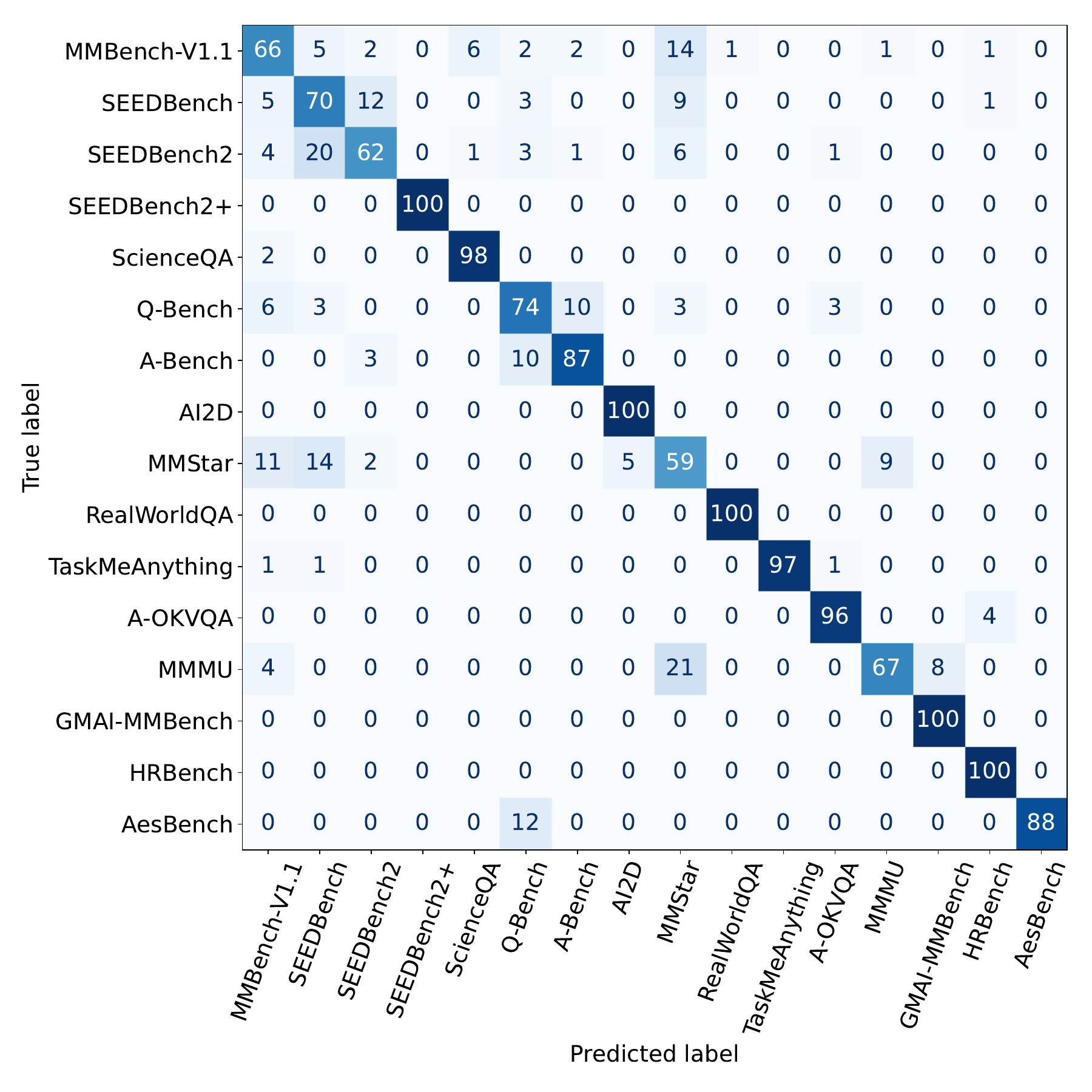} \\
         (e) Question + Answer & (f) Image + Question + Answer \\
    \end{tabular}
    \caption{The confusion matrices for MCQ with various inputs. The confusion matrix for MCQ using images and questions is shown in the Sec. 3.}
    \label{fig:mcq-cms}
\end{figure*}

\begin{figure*}
    \centering
    \begin{tabular}{cc}
         \includegraphics[width=0.4\linewidth]{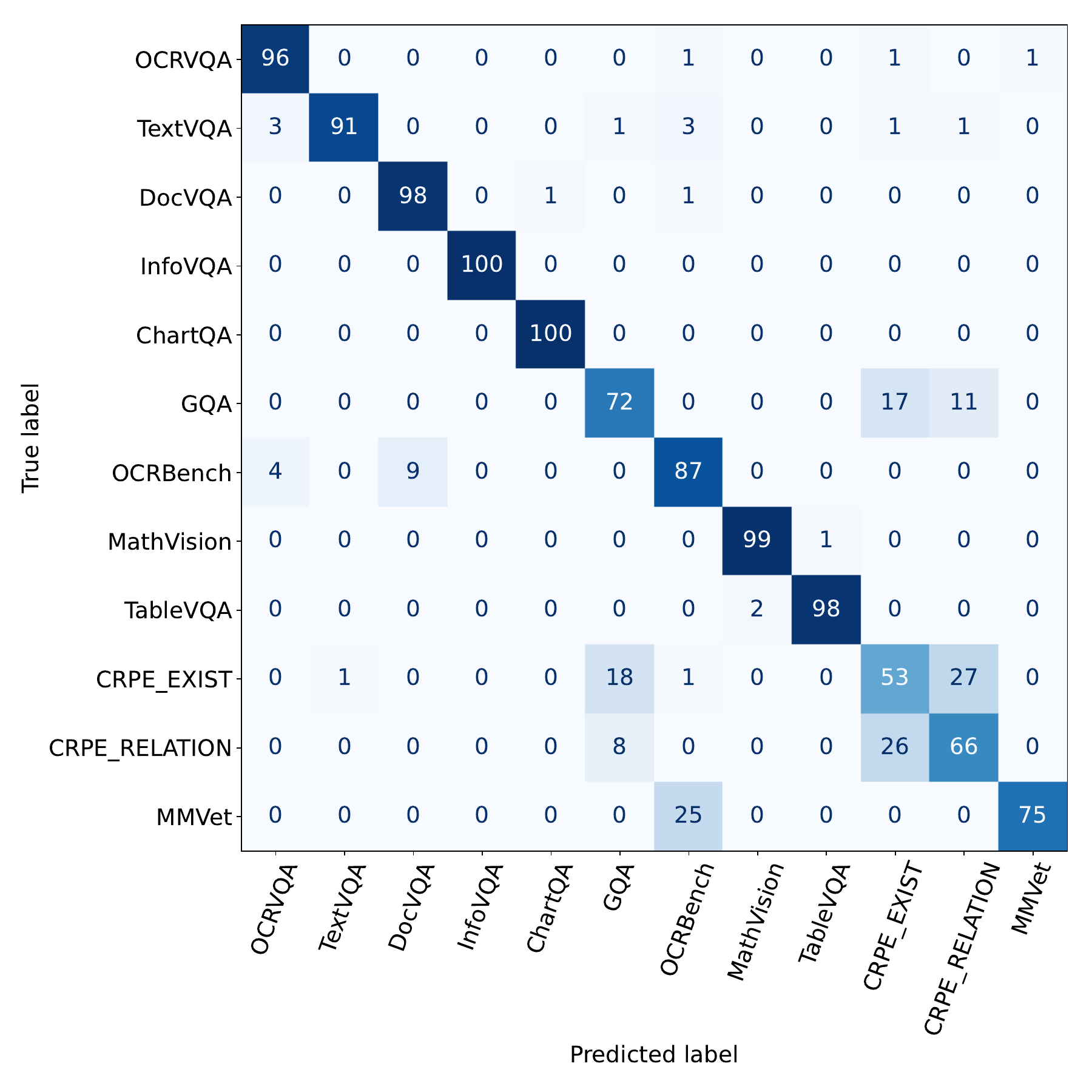} &
         \includegraphics[width=0.4\linewidth]{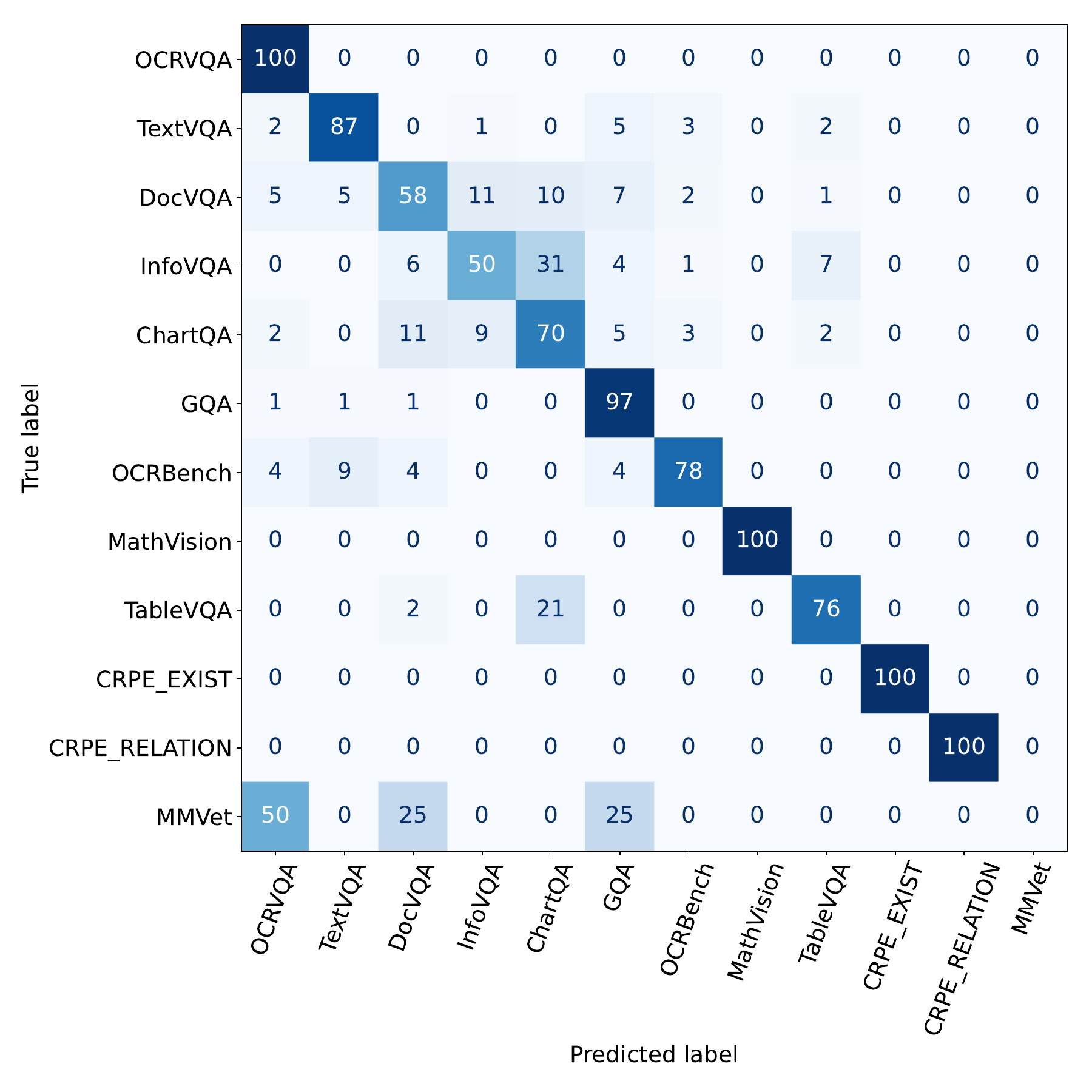} \\
         (a) Image & (b) Question \\
         \includegraphics[width=0.4\linewidth]{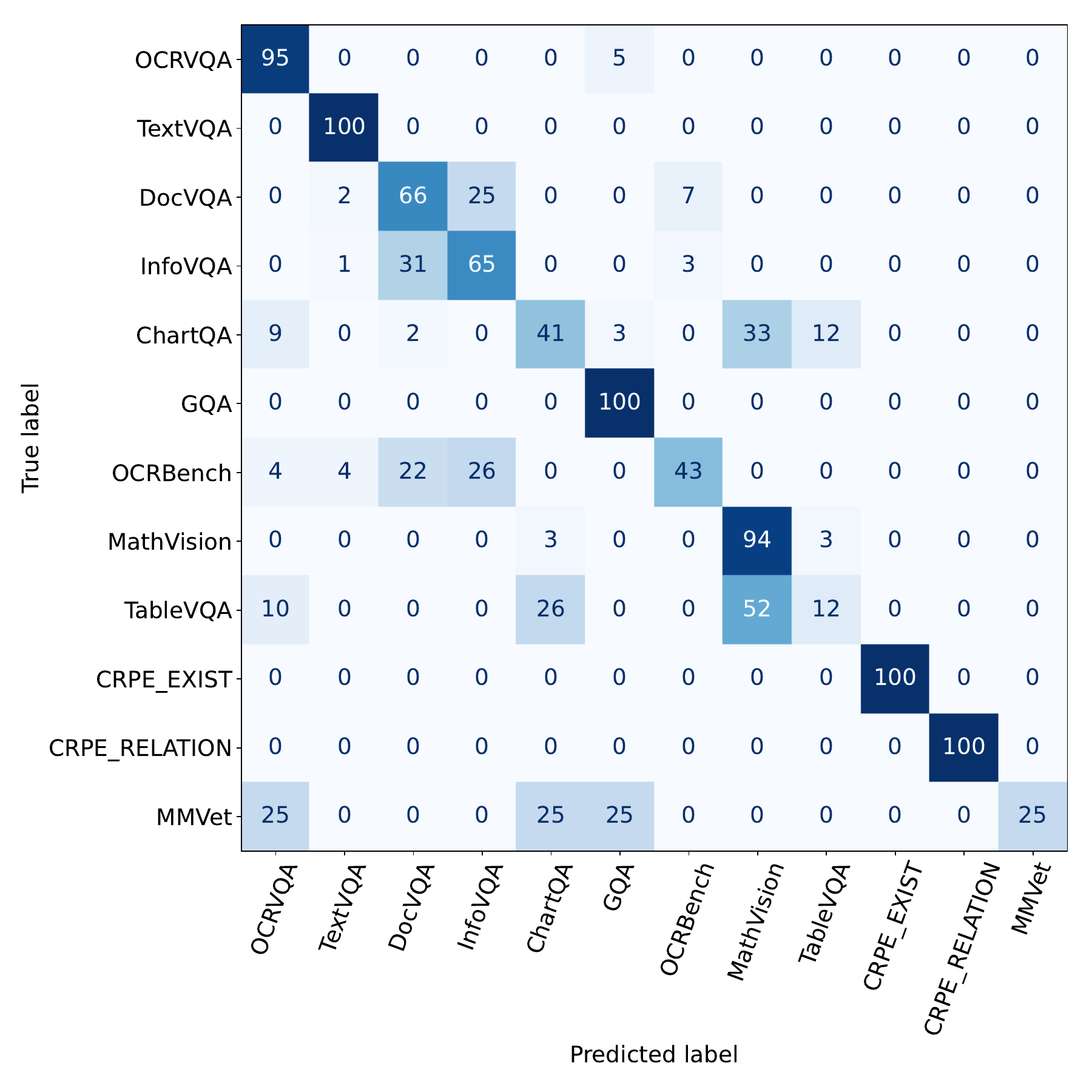} &
         \includegraphics[width=0.4\linewidth]{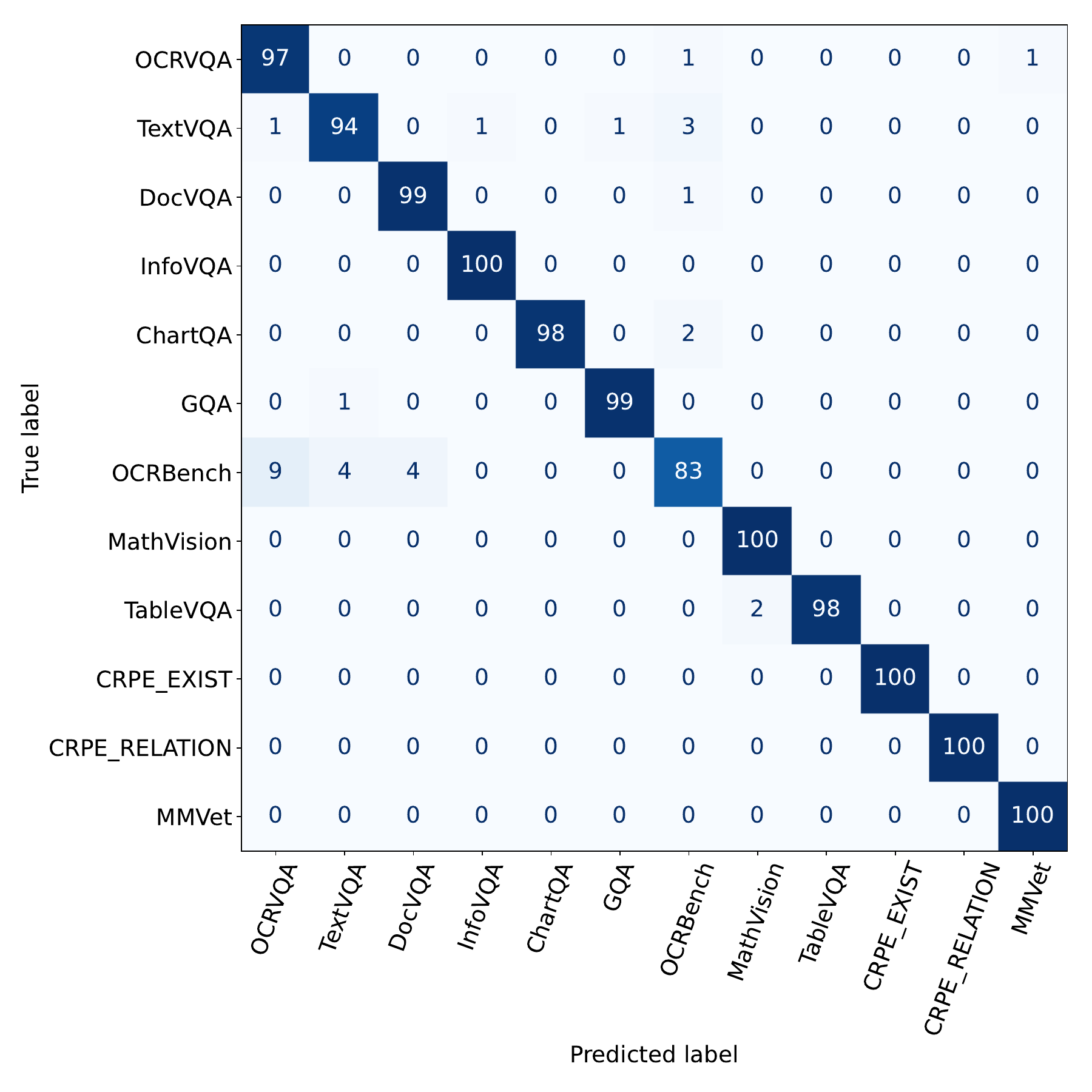} \\
         (c) Answer & (d) Image + Question \\
         \includegraphics[width=0.4\linewidth]{figures/confusion-matrix/vqa-confusion_matrix-ia.pdf} &
         \includegraphics[width=0.4\linewidth]{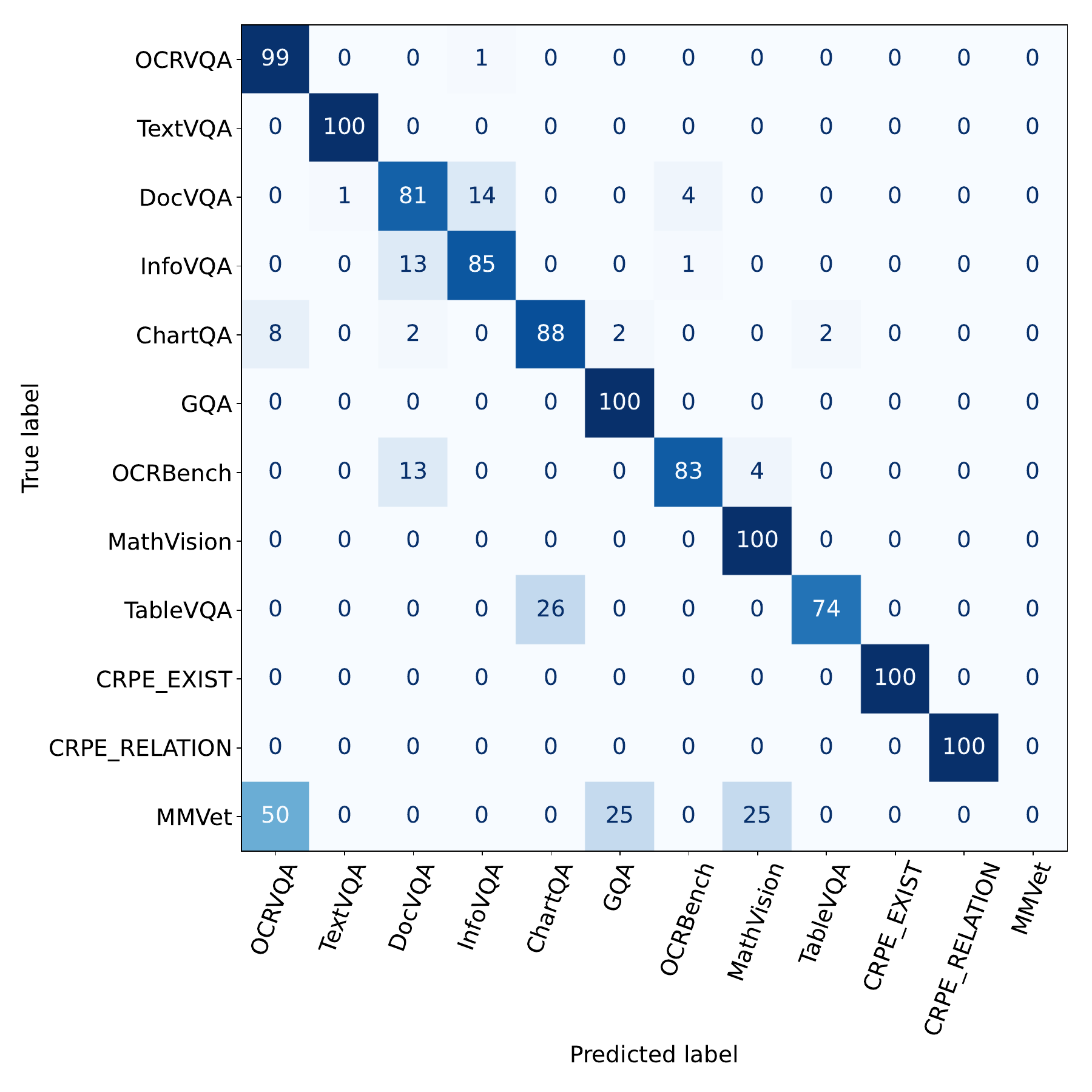} \\
         (e) Image + Answer & (f) Question + Answer \\
    \end{tabular}
    \caption{The confusion matrices for VQA with various inputs. The confusion matrix for VQA using images, questions, and answers is shown in the Sec. 3.}
    \label{fig:vqa-cms}
\end{figure*}

We show the confusion matrices of MCQ and VQA tasks with various inputs in Figs. \ref{fig:mcq-cms} and \ref{fig:vqa-cms}.
Since the answers in MCQ tasks lack meaningful content, predictions based solely on answers are essentially random.
In contrast, the other settings yield more meaningful results.

\section{Averaged Ranks of VLMs}
We show the $\mathrm{AvgRank}$ of VLMs, where the ranks of each model are averaged over the benchmarks.
We use following benchmarks for computing the $\mathrm{AvgRank}$: MMBench~\cite{MMBench}, SEEDBench~\cite{li2023seed}, SEEDBench2+~\cite{li2024seed2plus}, ScienceQA~\cite{lu2022learn}, Q-Bench~\cite{wu2023q}, A-Bench~\cite{zhang2024abench}, AI2D~\cite{Kembhavi2016ADI}, MMStar~\cite{chen2024we}, RealWorldQA~\cite{realworldqa}, TaskMeAnything~\cite{zhang2024task}, A-OKVQA~\cite{schwenk2022okvqa}, MMMU~\cite{yue2023mmmu}, GMAI-MMBench~\cite{chen2024gmai}, AesBench~\cite{huang2024aesbench}, OCRVQA~\cite{mishra2019ocr}, TextVQA~\cite{singh2019towards}, DocVQA~\cite{mathew2021docvqa}, InfographicVQA~\cite{mathew2022infographicvqa}, ChartQA~\cite{masry2022chartqa}, GQA~\cite{hudson2019gqa}, OCRBench~\cite{liu2023hidden}, MATH-Vision~\cite{wang2024measuring}, CRPE\_RELATION~\cite{wang2024allseeing_v2}, and MM-Vet~\cite{yu2023mm}

\begin{table}[hb]
  \centering
  \scalebox{0.9}{
  \begin{tabular}{lc}
    \toprule
        Model & Rank \\ 
        \midrule
        Qwen2-VL-7B-Instruct~\cite{wang2024qwen2} & 3.615 \\
        InternVL2-8B~\cite{chen2024far} & 5.538 \\
        InternVL-Chat-V1-5~\cite{chen2023internvl} & 5.846 \\
        glm-4v-9b~\cite{glm-4v} & 6.462 \\
        Ovis1.6-Llama3.2-3B~\cite{lu2024ovis} & 7.385 \\
        InternVL-Chat-V1-2-Plus~\cite{chen2023internvl} & 7.846 \\
        WeMM~\cite{wemm} & 8.269 \\
        MiniCPM-Llama3-V-2.5~\cite{yao2024minicpm} & 9.730 \\
        Qwen2-VL-2B-Instruct~\cite{wang2024qwen2} & 10.12 \\
        XComposer2~\cite{internlmxcomposer2} & 10.62 \\
        XinYuan-VL-2B-Instruct~\cite{xinyuan-vl} & 12.42 \\
        h2ovl-mississippi-2b~\cite{h2ovl} & 13.08 \\
        xgen-mm-phi3-dpo-r-v1.5~\cite{blip3-xgenmm} & 13.15 \\
        Phi-3.5-Vision~\cite{abdin2024phi} & 13.19 \\
        llava-next-llama3~\cite{llava-next} & 14.42 \\
        minimonkey~\cite{huang2024mini} & 15.12 \\
        InternVL2-2B~\cite{chen2024far} & 15.23 \\
        llava-next-vicuna-13b~\cite{llava-next} & 16.92 \\
        llava-next-interleave-7b~\cite{llava-next} & 17.00 \\
        llava-next-mistral-7b~\cite{llava-next} & 18.00 \\
        monkey-chat~\cite{li2023monkey} & 18.31 \\
        llava-next-vicuna-7b~\cite{llava-next} & 19.81 \\
        InternVL2-1B~\cite{chen2024far} & 19.85 \\
        monkey~\cite{li2023monkey} & 22.23 \\
        qwen-chat~\cite{bai2023qwen} & 22.46 \\
        qwen-base~\cite{bai2023qwen} & 24.42 \\
        idefics-9b-instruct~\cite{laurenccon2024obelics} & 25.96 \\
        \bottomrule
  \end{tabular}
  }
  \caption{$\mathrm{AvgRank}$ of VLMs.}
  \label{tab:AvgRank}
\end{table}

\section{Statistics of the Number of Parameters}
\begin{figure}
    \centering
    \includegraphics[width=1\linewidth]{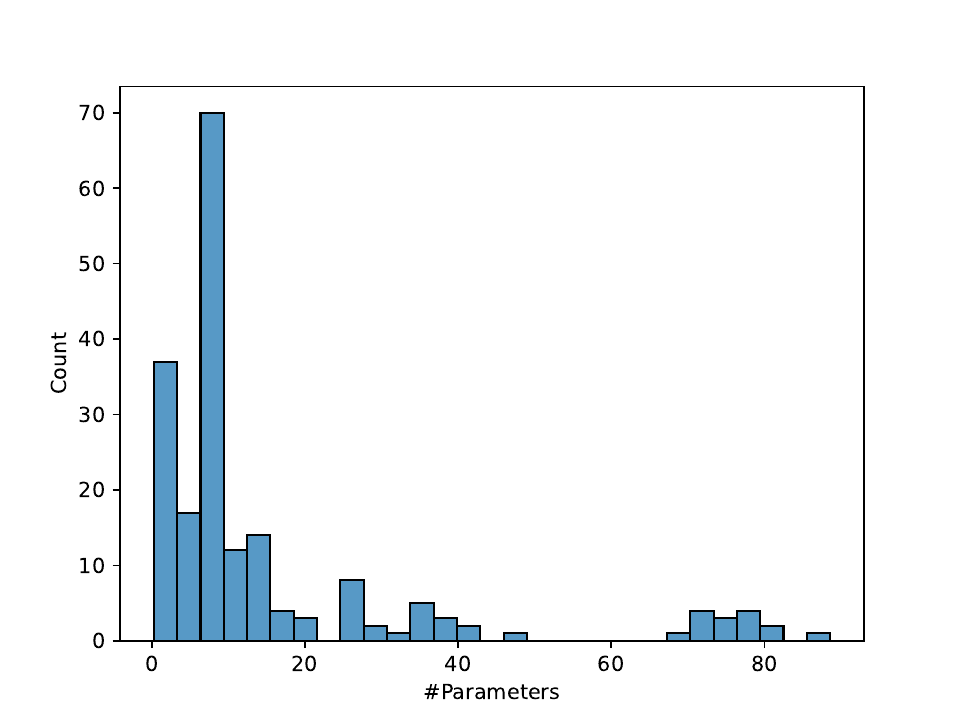}
    \caption{The histogram of the number of parameters of publicly available VLMs. We plot VLMs listed in Open VLM Leaderboard~\cite{vlm-lb}.}
    \label{fig:vlm-params}
\end{figure}

We show the histogram of the number of parameters of publicly available VLMs in Fig. \ref{fig:vlm-params}.
The figure shows that models with less than 15B parameters occupy 76.29\%.

\section{Component of ResampledBench-1000}
\begin{figure}
    \centering
    \includegraphics[width=1\linewidth]{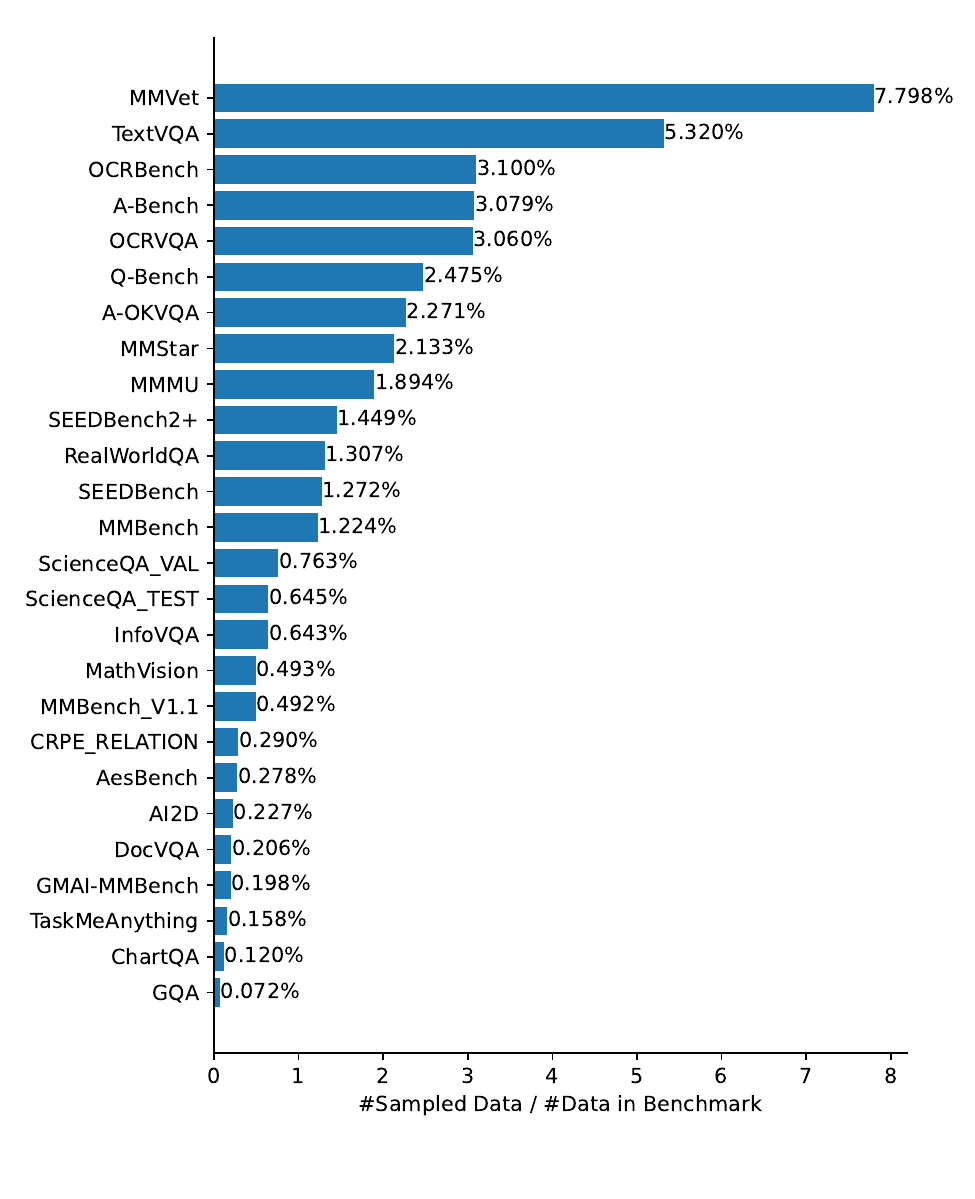}
    \caption{The ratio of sampled data for 1,000 samples to the original dataset size of the benchmarks.}
    \label{fig:bench-stats}
\end{figure}
We show the ratio of sampled data for 1,000 samples to the original dataset size of the benchmarks in Fig. \ref{fig:bench-stats}.
A relatively large proportion of data has been sampled from MMVet and TextVQA, which implies that they are relatively comprehensive compered to the other benchmarks.
Interestingly, the correlation with the average rank shown in Fig. \ref{fig:corr_rank} does not necessarily indicate how likely a dataset is to be sampled.
This is because a low correlation only suggests a lack of comprehensiveness in the data, but it does not imply that the dataset contains fewer samples useful for evaluation.

\end{document}